\documentclass[letterpaper]{article} 
\usepackage{aaai2026}  
\usepackage{times}  
\usepackage{helvet}  
\usepackage{courier}  
\usepackage[hyphens]{url}  
\usepackage{graphicx} 
\usepackage{natbib}  
\usepackage{caption} 
\usepackage{algorithm}
\usepackage{algorithmic}

\usepackage[percent]{overpic}
\usepackage{multirow}
\usepackage{pifont} 
\usepackage{enumitem}
\usepackage{amsmath}
\usepackage{amssymb}
\usepackage{booktabs}
\usepackage{xcolor}

\usepackage{newfloat}
\usepackage{listings}
\DeclareCaptionStyle{ruled}{labelfont=normalfont,labelsep=colon,strut=off} 
\floatstyle{ruled}
\newfloat{listing}{tb}{lst}{}
\floatname{listing}{Listing}
\title{Axis-Aligned Document Dewarping}
\author{
    Chaoyun Wang\textsuperscript{\rm 1}, 
    I-Chao Shen\textsuperscript{\rm 2}, 
    Takeo Igarashi\textsuperscript{\rm 2},  
    Caigui Jiang\textsuperscript{\rm 1}\thanks{Corresponding Author.}
}
\affiliations{
    \textsuperscript{\rm 1}National Key Laboratory of Human-Machine Hybrid Augmented Intelligence, National Engineering Research Center of Visual Information and Applications, and Institute of Artificial Intelligence and Robotics, Xi'an Jiaotong University\\
    \textsuperscript{\rm 2}The University of Tokyo\\
chaoyunwang@stu.xjtu.edu.cn, jdilyshen@gmail.com, takeo@acm.org, cgjiang@xjtu.edu.cn
}

\usepackage{bibentry}
\usepackage{cleveref}

\defcitealias{verhoeven2023uvdoc}{Verhoeven et al. 2023}

\begin{document}

\maketitle

\begin{abstract}
Document dewarping is crucial for many applications. However, existing learning-based methods rely heavily on supervised regression with annotated data without fully leveraging the inherent geometric properties of physical documents.
Our key insight is that a well-dewarped document is defined by its axis-aligned feature lines. This property aligns with the inherent axis-aligned nature of the discrete grid geometry in planar documents. 
Harnessing this property, we introduce three synergistic contributions: for the training phase, we propose an axis-aligned geometric constraint to enhance document dewarping; for the inference phase, we propose an axis alignment preprocessing strategy to reduce the dewarping difficulty; and for the evaluation phase, we introduce a new metric, Axis-Aligned Distortion (AAD), that not only incorporates geometric meaning and aligns with human visual perception but also demonstrates greater robustness.
As a result, our method achieves state-of-the-art performance on multiple existing benchmarks, improving the AAD metric by 18.2\% to 34.5\%. The code is publicly available at https://github.com/chaoyunwang/AADD.
\end{abstract}

\section{Introduction} \label{sec:Introduction}
The digitization of documents dramatically facilitates our lives by converting information on physical documents into electronic data that can be used with digital devices. However, traditional scanners, which require fixed positioning for precise acquisition, have limitations when dealing with paper deformation. With the widespread use of mobile phones and cameras, photography has become a primary tool for digitization. Nonetheless, images captured by handheld cameras often suffer from geometric distortions and uneven lighting, which degrade the reading experience and hinder downstream tasks like optical character recognition (OCR).

In early research, many studies defined this problem as surface reconstruction from different imaging devices, including multi-view images~\cite{tsoi2007multi,liang2008geometric}, stereo cameras~\cite{ulges2004document}, and structured light cameras~\cite{brown2001document,meng2014active}. Subsequent studies on single-view images have leveraged various document priors, such as document boundaries~\cite{brown2006geometric}, text lines~\cite{kil2017robust,kim2015document}. However, the performance of these methods is often limited by the robustness of low-level feature detection.

With the recent development of deep learning, convolutional neural networks began to learn 2D deformation field mapping from deformed images to expected dewarping results~\cite{ma2018docunet,das2019dewarpnet,xie2020dewarping,feng2021doctr,ma2022learning,li2023layout,li2023foreground,feng2022geometric,jiang2022revisiting,yu2024docreal}. Data-driven methods perform supervised learning dewarping under the guidance of various supervision signals with synthetic annotations~\cite{ma2018docunet,li2023layout,verhoeven2023uvdoc},
such as depth~\cite{das2019dewarpnet}, geometric semantics~\cite{markovitz2020can}, control points~\cite{verhoeven2023uvdoc, xie2021document,yu2024docreal}, layout~\cite{li2023layout}.

Although prior methods perform well on simple deformations, they often fail on document images with complex deformations. This is mainly because previous studies rely on strong supervision signals, such as control points, which are effective but have no geometric meaning, and text lines, which are meaningful but difficult to extract and often exhibit poor generalization. These methods often overlook the fundamental geometric properties of paper documents—a gap our work aims to fill.
\begin{figure}[!htbp]
    \centering
    \includegraphics[width=1.0\linewidth]{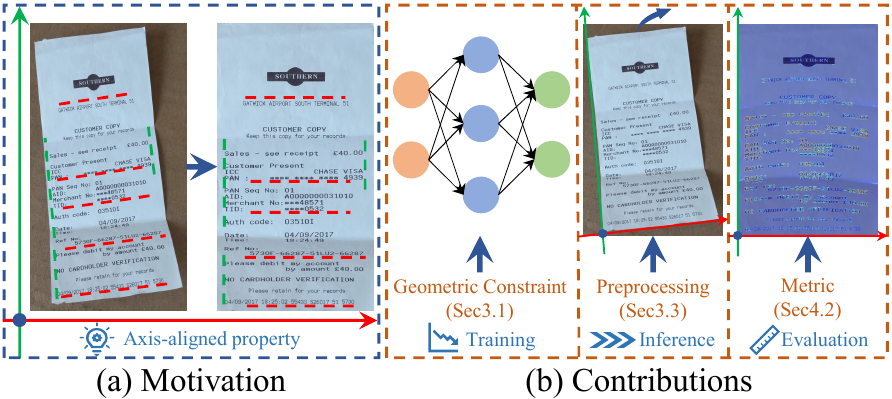}
    \caption{
    Our research motivation and main contributions. (a) A warped document (left) and its target rectified version (right). The key characteristic of the rectified version is the alignment of its feature lines (highlighted by dotted lines) with the axes. We term this the ``axis-aligned property". (b) Inspired by (a), we integrate this axis-aligned property into the training, inference, and evaluation stages of our deep learning method.}
    \label{fig:intro_img}
\end{figure}

In this paper, we perform document dewarping using a lightweight convolutional neural network that predicts a 2D unwarping grid from the distorted image. Our key insight, as illustrated in \Cref{fig:intro_img}(a), is that the defining characteristic of a successfully rectified document is the alignment of its feature lines with the coordinate axes. We term this fundamental observation the ``axis-aligned property"—a principle rooted in the inherent discrete grid geometry of planar documents.

Inspired by this single, powerful principle, we systematically integrate it into all three key stages of our deep learning pipeline, as outlined in \Cref{fig:intro_img}(b):
For training, we design an axis-aligned geometric constraint that explicitly guides the network to learn this property, significantly improving dewarping performance. For inference, we introduce a novel axis-alignment preprocessing strategy. This strategy leverages the property to normalize the document's orientation beforehand, simplifying the task for the network and boosting its robustness against various rotations and scales. Finally, for evaluation, we propose a new metric, Axis-Aligned Distortion (AAD), which directly quantifies how well the dewarped result adheres to the axis-aligned property, offering a measure that is both geometrically meaningful and consistent with human perception.

Our contributions are as follows.
\begin{itemize}[leftmargin=2em]
    \item We introduce an axis-aligned geometric constraint, significantly enhancing the performance of learning-based document dewarping.
    \item We propose an axis alignment preprocessing strategy in the inference phase which boosts the dewarping performance robustly.
    \item We propose a new evaluation metric, AAD, that not only incorporates geometric meaning and aligns with human visual perception but also demonstrates enhanced robustness.
\end{itemize}

\section{Related Work} \label{sec:related_work}
Current research on document dewarping can be broadly categorized into traditional optimization methods and deep learning-based methods.
As mentioned in~\Cref{sec:Introduction}, traditional approaches to document dewarping primarily involve image processing and 3D reconstruction, preceding the advent of deep learning.

Research in image processing has focused on modeling the most commonly used low-level features extracted from document images, including letters/words, boundaries, spacing, and text lines, and to correct perspective and page curl distortion through local line spacing and cell shape, which can be formulated as an energy minimization optimization problem~\cite{ulges2005document,stamatopoulos2010goal,kim2015document,takezawa2017robust}. More recently, Jiang et al.~\cite{jiang2022revisiting} applied deep learning to extract boundaries and document lines, followed by geometric constraint optimization.
However, reliably detecting such low-level features in heavily distorted images is challenging, limiting the applicability of these methods.

In 3D reconstruction research, variations in shadows from a single image or across different imaging devices, along with other input cues, have been utilized to reconstruct 3D surfaces~\cite{tan2005restoring,wada1997shape,zhang2009unified,liang2008geometric,he2013book}.
Some studies model these surfaces as special types, such as cylindrical surfaces~\cite{cao2003cylindrical} and developable surfaces~\cite{meng2014active}. Recently, Luo et al.~\cite{luo2022geometric} treat the document as a developable surface and employ a recently proposed isometric constraint for modeling~\cite{jiang2020quad}. However, these methods rely on detecting document feature lines, resulting in long optimization times and limited generality.

Deep learning methods leverage large amounts of data generated by rendering techniques and dataset collection, such as Doc3D~\cite{das2019dewarpnet}, Inv3D~\cite{Hertlein2023}, Simulated paper~\cite{li2023layout}, UVDoc~\citepalias{verhoeven2023uvdoc}.
These data-driven approaches can be categorized into forward mapping methods~\cite{ma2018docunet,li2019document,jiang2022revisiting} and backward mapping methods~\cite{das2019dewarpnet,das2021end,feng2021doctr,feng2022geometric}.
To further improve optimization, some methods incorporate additional supervision signals, such as text lines, masks, boundaries, layouts, angles, and curvatures to further improve optimization~\cite{zhang2022marior,feng2021doctr,dai2023matadoc,li2023foreground,li2023layout}. However, extracting these additional cues can introduce extra noise and lack generalizability.

An effective approach is to fully exploit the inherent geometric properties of the document, which are independent of annotated labels and generally more robust.
As Xie et al.~\cite{xie2020dewarping} exploited the continuity property of document deformation by incorporating local fairing constraints into the displacement prediction process.
Furthermore, the recent work~\cite{wang2024gso,wang2025interactive}, based on grid surface modeling, achieves accurate and robust optimization by learning the intrinsic geometric properties.

In this paper, we study and leverage the intrinsic geometric properties for document dewarping.

\section{Method} \label{sec:Method}
We adopt the convolutional neural network architecture from UVDoc~\citepalias{verhoeven2023uvdoc} as the foundation for our document dewarping network. This fully convolutional deep neural network simultaneously predicts the document’s 3D grid mesh and corresponding 2D unwarping grid in a dual-task framework, which is inspired by the work of Xie et al.~\cite{xie2020dewarping,xie2021document}

In the following sections,~\Cref{sec:slgc} details the axis-aligned geometric constraint introduced during training,~\Cref{sec:loss} analyzes the loss functions employed during network training, and~\Cref{sec:rotation_alignment} describes the preprocessing strategy for axis alignment during inference.

\subsection{Axis-Aligned Geometric Constraints} \label{sec:slgc}

\begin{figure}[!htbp]
    \centering
    \includegraphics[width=1.0\linewidth]{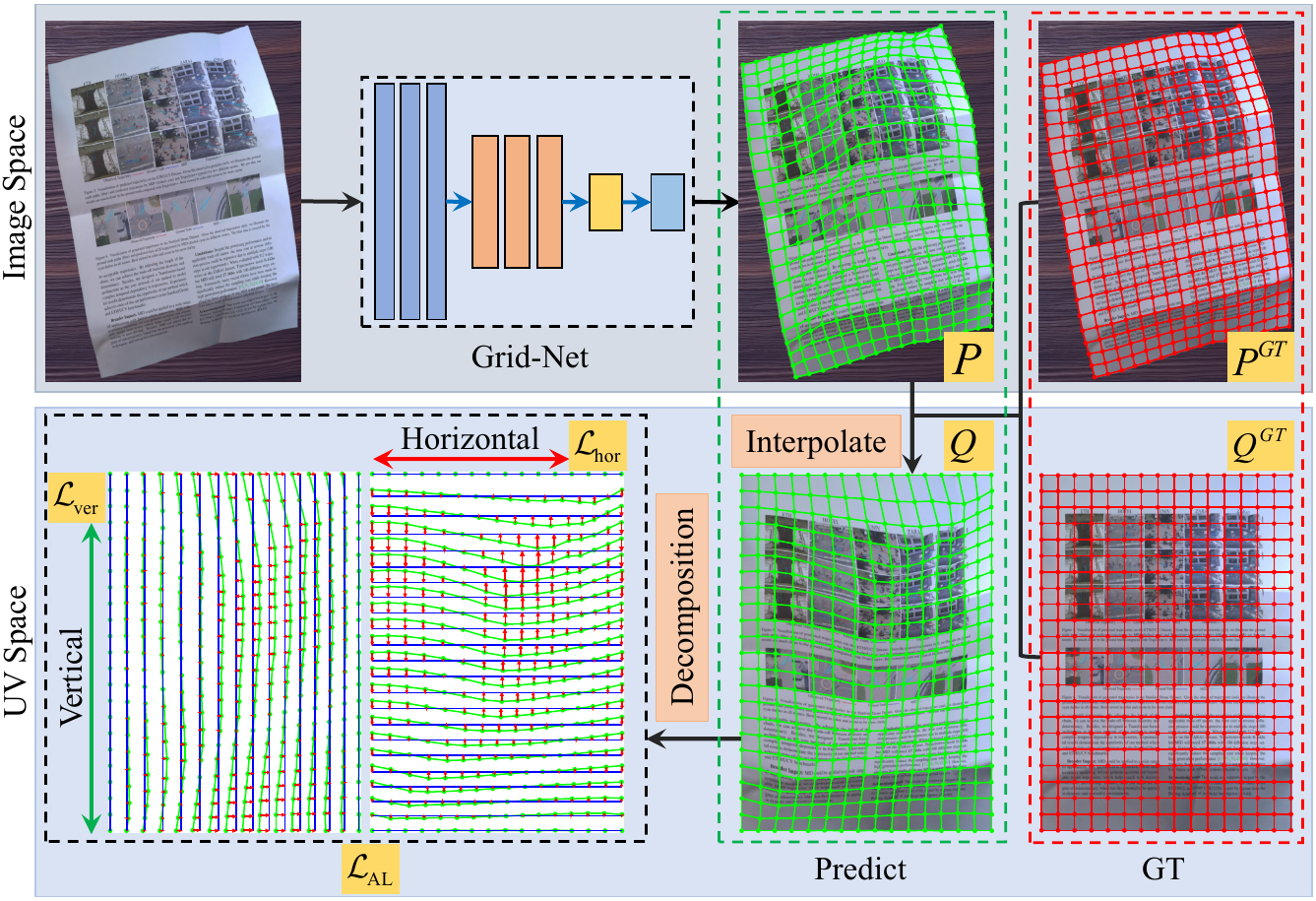}
      \caption{Axis-aligned geometric constraints were enforced during document dewarping network training. The top row shows the predicted 2D unwarping grid (obtained from Grid-Net in UVDoc~\citepalias{verhoeven2023uvdoc}), and the bottom row displays its transformation into UV space based on the ground truth, facilitating the computation of the corresponding axis-aligned geometric constraints loss along horizontal and vertical directions.}
    \label{fig:pipeline_uvstraight}
\end{figure}

\Cref{fig:pipeline_uvstraight} illustrates the pipeline for computing the axis-aligned geometric constraint loss during training. Our approach takes advantage of the inherent geometric properties of planar paper, where the corresponding discrete grid geometry is aligned along the horizontal and vertical axes (as shown by \(Q^{GT}\) in~\Cref{fig:pipeline_uvstraight}). This property aligns with the horizontal and vertical orientations of the document’s feature lines in image space. Therefore, we can assess the axis alignment distortion of these feature lines in the dewarped image by computing the corresponding deviation of the unwrapped grid in planar space. However, since the current predictions are in image space, directly calculating this error is challenging.

To address this issue, we first employ an interpolation function to map the predicted 2D unwarping grid from image space \(P\) to UV space \(Q\). In the UV space, the deviation from horizontal and vertical alignment is effectively quantified by computing the position variance along each row and column. By integrating these axis-aligned geometric constraints into the training process, we can guide the network in learning to dewarp distorted document feature lines toward the axis-aligned direction.

Given a distorted document image, a deep convolutional network predicts a 2D unwrapped grid points with sizes \(h \times w \times 2\). Each point on the grid can be represented as:
\begin{equation}
P = \{p_{i,j}\}_{i=1,j=1}^{h, w}, \quad p_{i,j} = (x_{i,j}, y_{i,j}).
\label{eq:predicted_points_grid}
\end{equation}
During training, the corresponding ground truth grid is:
\begin{equation}
P^{GT} = \{p^{GT}_{i,j}\}_{i=1,j=1}^{h, w}, \quad p^{GT}_{i,j} = (x^{GT}_{i,j}, y^{GT}_{i,j}).
\label{eq:gt_points_grid}
\end{equation}
In UV space, these ground truth grid are a uniform grid:
\begin{equation}
Q^{GT} = \{q^{GT}_{i,j}\}_{i=1,j=1}^{h, w}, \quad q^{GT}_{i,j} = (u^{GT}_{i,j}, v^{GT}_{i,j}).
\label{eq:gt_uv_grid}
\end{equation}

To quantify the distortion of the prediction results, we distort the correspondence points in \(Q^{GT}\) according to the relative grid‐position relationships in image space:
\begin{equation}
  q = f(p) = \mathrm{Interpolate}\bigl(q^{GT};\,p^{GT},\,P\bigr).
  \label{eq:interpolation}
\end{equation}
Thus, each predicted 2D grid point is transformed into the UV space as follows:
\begin{equation}
Q = \{q_{i,j}\}_{i=1,j=1}^{h, w}, \quad q_{i,j} = f(p_{i,j})=(u_{i,j}, v_{i,j}).
\label{eq:mapped_points_grid}
\end{equation}
In UV space, the axis-aligned error of 2D grid \(Q\) consists of two components.
The horizontal error is quantified by the variance of the \(v\) values along each row, and the vertical error by the variance of the \(u\) values along each column, computed as follows:
\begin{equation}
\begin{aligned}
\mathcal{L}_{\text{hor}} &= \sum\nolimits_{j=1}^{h} \mathrm{Var}\Big(\{v_{1,j}, v_{2,j}, \dots, v_{w,j}\}\Big), \\
\mathcal{L}_{\text{ver}} &= \sum\nolimits_{i=1}^{w} \mathrm{Var}\Big(\{u_{i,1}, u_{i,2}, \dots, u_{i,h}\}\Big).
\end{aligned}
\label{eq:loss_combined_grid}
\end{equation}

The overall axis-aligned geometric constraint loss is then defined as:
\begin{equation}
\mathcal{L}_{\text{AL}} = \mathcal{L}_{\text{hor}} + \mathcal{L}_{\text{ver}}.
\label{eq:total_loss_grid}
\end{equation}

Minimizing \(\mathcal{L}_{\text{AL}}\) encourages the predicted 2D unwarping grid, when mapped to the UV space, to align with the axes as closely as possible.

\subsection{Loss Function}\label{sec:loss}

Let \(I \in \mathbb{R}^{H \times W \times 3}\) be the input image. The network predicts 2D unwarping grid \(G_{2D} \in \mathbb{R}^{h \times w \times 2}\) and corresponding 3D grid mesh \(G_{3D} \in \mathbb{R}^{h \times w \times 3}\). The dewarped image \(I_d\) is obtained via an interpolation and grid sampling using \(I\) and \(G_{2D}\).
Our overall training loss is composed of the following components:

\textbf{Grid 2D and 3D Losses.} 
Following UVDoc~\citepalias{verhoeven2023uvdoc}, we supervise the network with the losses:
\begin{equation}
\mathcal{L}_{2D} = \| G_{2D} - G_{2D}^{\text{gt}} \|_1, \mathcal{L}_{3D} = \| G_{3D} - G_{3D}^{\text{gt}} \|_1.
\label{eq:loss2d-3d}
\end{equation}
where \(G_{2D}^{\text{gt}}\) and \(G_{3D}^{\text{gt}}\) denote the ground truth 2D grid and 3D coordinates, respectively.

\textbf{Axis-Aligned Geometric Constraint Loss.} 
We impose axis-aligned geometric constraints \(\mathcal{L}_{AL}\) in the UV space, guiding the network to achieve effective axis alignment dewarping of feature lines, as depicted in~\Cref{sec:slgc}.

\textbf{SSIM Loss.} 
Since geometric dewarping may introduce artifacts (e.g., shadows) in the dewarped images, a pixel-level MSE loss might over-constrain model optimization and lead to instability. Hence, we incorporate an SSIM (Structural Similarity Index Measure) loss to measure the similarity between the dewarped image and the ground truth, capturing both local structure and global consistency, which helps the model converge effectively:
\begin{equation}
\mathcal{L}_{\text{SSIM}} = 1 - \mathrm{SSIM}(I, I_d).
\label{eq:lossssim}
\end{equation}

The overall loss is defined as:
\begin{equation}
\mathcal{L}_{\text{all}} = \alpha\,\mathcal{L}_{2D} + \beta\,\mathcal{L}_{3D}  + \gamma\,\mathcal{L}_{AL} + \lambda\,\mathcal{L}_{\text{SSIM}}.
\label{eq:lossall}
\end{equation}
with hyperparameters set as \(\alpha = \beta=1, \gamma =0.2, \lambda= 0.05\), in our experiments.

\subsection{Axis Alignment Preprocessing for Inference}\label{sec:rotation_alignment}
During inference, previous methods employ additional segmentation and deformation models for preprocessing~\cite{ma2022learning,feng2021doctr} to improve dewarping robustness under diverse input conditions. Essentially, these methods distinguish between the background and the document region, enabling the model to focus on the document area and reducing dewarping difficulty. However, they do not directly address the elimination of geometric distortion.

\begin{figure}[!htbp]
    \centering
    \includegraphics[width=1.0\linewidth]{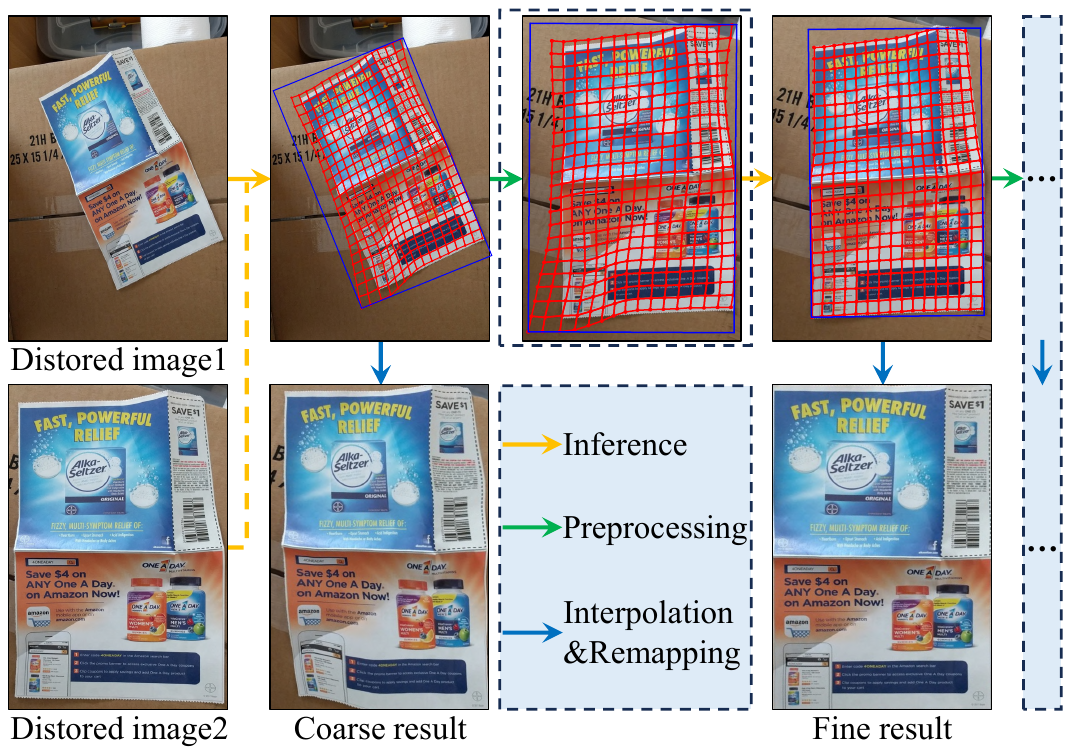}
    \caption{Illustration of the dewarping inference process with axis alignment preprocessing. The red grid represents the predicted 2D unwarping grid, while the blue rectangle indicates the minimum-area rotated rectangle computed from it.}
    \label{fig:rotate_aligned}
\end{figure}

Drawing on the insights from~\Cref{fig:intro_img}(a), axis alignment is central to the dewarping process. So, we propose an axis alignment preprocessing strategy, as illustrated in~\Cref{fig:rotate_aligned}. For a given input image, we first compute the minimum‐area rotation rectangle using the positional information from the predicted 2D unwarping grid. Subsequently, we rotate the image to align the document's principal axes and clip the target area. This preprocessed image is then fed back into the network for a refined prediction.
If necessary, this process can be iteratively repeated to achieve further improvement under varying conditions. As shown in~\Cref{fig:rotate_aligned}, the fine results obtained after preprocessing outperform the coarse results, effectively reducing the influence of the target position on the overall correction.

Unlike prior methods that rely on external segmentation models~\cite{ma2022learning,feng2021doctr}, our preprocessing is self-contained. It directly leverages the predicted 2D unwarping grid to extract object region information, rendering the method more efficient and elegant. A detailed analysis is provided in~\Cref{sec:ablation study}.

\section{Experiments} \label{sec:Experiemnts}
The experimental section introduces the training data, test benchmarks, evaluation metrics, and implementation details, subsequently, we analyze the experimental results both quantitatively and qualitatively.

\subsection{Datasets and Benchmarks}
\subsubsection{Training Datasets}
We used two datasets, Doc3D~\cite{das2019dewarpnet} and UVDoc~\citepalias{verhoeven2023uvdoc}, for alternative optimization models following the setting of~\citepalias{verhoeven2023uvdoc}.

\textbf{Doc3D}~\cite{das2019dewarpnet} is the largest distortion dewarping dataset, containing 100K distorted images. It was created using approximately 4,000 authentic document images and rendering software. The labels for each distorted document image include 3D coordinate maps, albedo maps, normals, depth maps, UV maps, and backward mapping maps.

\textbf{UVDoc}~\citepalias{verhoeven2023uvdoc} aims to reduce the domain gap between the synthetic Doc3D dataset~\cite{das2019dewarpnet}—commonly used for training document unwarping models—and real document photographs. This dataset contains 20K pseudo-photorealistic document images and provides all the necessary information to train a coarse grid-based document unwarping neural network.

\subsubsection{Benchmarks}
We evaluated our method on the following two benchmarks:

\textbf{DocUNet}~\cite{ma2018docunet}: This benchmark comprises 130 distorted images captured in natural scenes using mobile devices. Following previous methods, 25 rich-text images were selected for OCR evaluation, and according to DocGeoNet~\cite{feng2022geometric}, the 127th and 128th images (which did not match the labels) were rotated by 180$^\circ$.

\textbf{DIR300}~\cite{feng2022geometric}: Captured with a moving camera, this benchmark contains 300 images of real distorted documents with complex backgrounds, varying degrees of distortion, and diverse lighting conditions. In line with previous studies, 90 rich-text images were selected for OCR evaluation.

\subsection{Evaluation Metrics}
We evaluate our approach using image similarity metrics and optical character recognition (OCR) performance.

\textbf{MS-SSIM, LD and AD.}
Following Ma et al.~\cite{ma2018docunet}, we employ multi-scale structural similarity (MS-SSIM) and local distortion (LD) to assess image similarity. Additionally, we use the aligned distortion (AD) metric introduced in~\cite{ma2022learning} to overcome the limitations of earlier measures. 
 
\textbf{CER and ED.}
Consistent with previous work, we measure OCR performance using Character Error Rate (CER)~\cite{morris2004and} and Edit Distance (ED)~\cite{levenshtein1966binary}. Our experiments utilize OCR recognition with Tesseract v5.4.0.20240606 and PyTesseract v0.3.13~\footnote{https://pypi.org/project/pytesseract/}.

Subsequently, we introduce the newly proposed metric.

\subsubsection{AAD Metric}\label{sec:AAD}
As illustrated in~\Cref{fig:intro_img}(a), our visual evaluation of dewarped documents primarily relies on the axis alignment error of the document's feature lines, which carries clear geometric significance. However, existing metrics fail to capture this characteristic effectively. 
We introduce the AAD (Axis-Aligned Distortion) metric, which quantifies distortion by measuring the axial alignment error of the dewarped results. 
It is conceptually identical to our geometric constraint (\Cref{sec:slgc}), with the key difference being its application: it is computed on image pairs during evaluation, rather than on 2D grid pairs during training.

Let \(y\) denote the ground truth image. The optical flow field from \( y \) to its dewarped version is computed using the SIFT-flow algorithm~\cite{liu2010sift}, yielding flow components \( v_x \) and \( v_y \) along the horizontal and vertical directions, respectively. The directional gradients of \( y \) are then extracted using the Sobel operator~\cite{sobel19683x3} along each axis and normalized:
\begin{equation}
    \tilde{g}_i = \frac{\left|\text{Sobel}_i(y)\right|}{\max\left(\left|\text{Sobel}_i(y)\right|\right)}, \quad i \in \{x, y\}.
\end{equation}

For each row \(i\) and column \(j\), we compute the gradient-weighted means and the corresponding deviations as follows:
\begin{equation}
\label{eq:deviation}
\begin{alignedat}{2}
&m_i = \frac{\sum_{j} v_y(i,j)\,\tilde{g}_y(i,j)}{\sum_{j} \tilde{g}_y(i,j) + \epsilon},  
d^{\text{row}}_{i,j} = \tilde{g}_y(i,j) \cdot \left| v_y(i,j) - m_i \right|, \\
&n_j = \frac{\sum_{i} v_x(i,j)\,\tilde{g}_x(i,j)}{\sum_{i} \tilde{g}_x(i,j) + \epsilon}, 
d^{\text{col}}_{i,j} = \tilde{g}_x(i,j) \cdot \left| v_x(i,j) - n_j \right|.
\end{alignedat}
\end{equation}
Here, \(m_i\) and \(n_j\) denote the gradient-weighted mean of the vertical flow in the \(i\)th row and the horizontal flow in the \(j\)th column, respectively, while \(d^{\text{row}}_{i,j}\) and \(d^{\text{col}}_{i,j}\) measure the corresponding deviations in pixel (\(i, j\)).

Define the pixel-wise overall deviation as the Euclidean combination of the row and column deviations, and let the AAD be their average over all \(N\) pixels:
\begin{equation}
\label{eq:AAD}
\begin{aligned}
d_{i,j} = \sqrt{\left[d^{\text{row}}_{i,j}\right]^2 + \left[d^{\text{col}}_{i,j}\right]^2}, 
\text{AAD} = \frac{1}{N} \sum_{i,j} d_{i,j}.
\end{aligned}
\end{equation}

The AAD metric uses the normalized image direction gradient weight to calculate the axial alignment error of the feature line in the row and column, lower values indicate a better dewarping effect.

\begin{figure}[!htbp]
    \centering
    \includegraphics[width=1.0\linewidth]{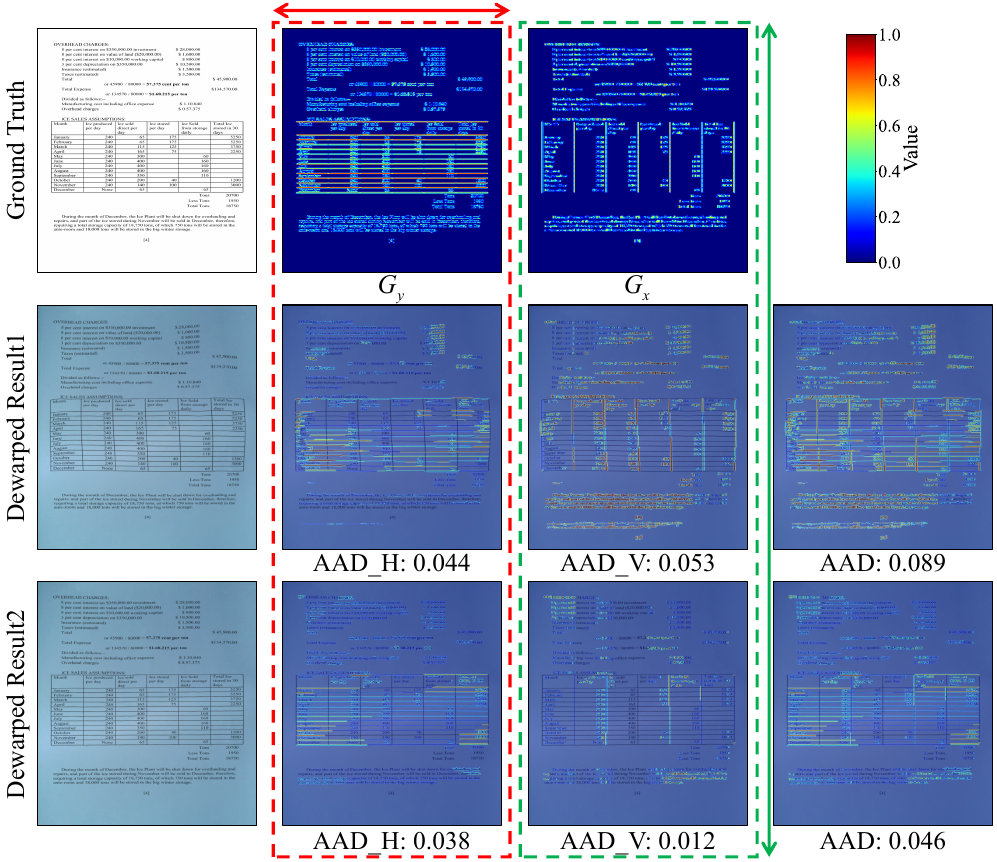}
    \caption{Example of visualizing AAD metrics. The first row is the ground truth image and its gradient heatmap in the horizontal ($G_{y}$) and vertical ($G_{x}$) directions; the second and third rows correspond to the two different dewarped results and their overlay heatmaps for AAD and its horizontal (AAD\_H) and vertical (AAD\_V) components, the $\leftrightarrow$ and $\updownarrow$ arrows indicate the directions for visualizing the axis-aligned error of the dewarped feature lines.}
    \label{fig:AAD}
\end{figure}

\Cref{fig:AAD} shows an example visualization of the AAD metric. For two different dewarped results, the alignment of table feature lines and the horizontal alignment of text lines serve as the primary evaluation criteria. By overlaying the horizontal (row) and vertical (col) components of the AAD heatmap onto the dewarped results, the corresponding axis alignment errors are displayed in detail. The heatmap colors correlate well with the calculated values, offering a perceptual measure of axis alignment.

\begin{figure}[!htbp]
    \centering
    \includegraphics[width=1.0\linewidth]{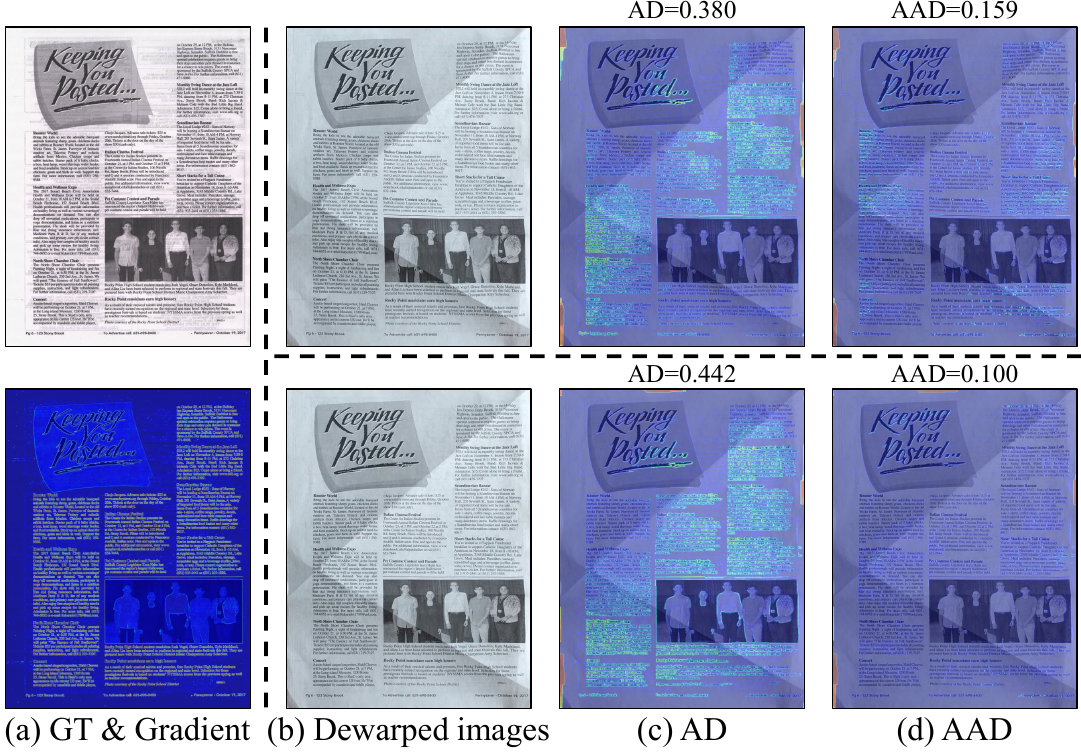}
    \caption{Visualization and comparison of the AD and AAD metrics on a example. (a) The ground truth image (top) and its corresponding gradient map (bottom). (b) Dewarped results from~\citepalias{verhoeven2023uvdoc} (top) and our method (bottom). (c) Heatmaps of the AD metric for the results shown in (b). (d) Corresponding heatmaps of our proposed AAD metric.}
    \label{fig:AD&ADD}
\end{figure}

\Cref{fig:AD&ADD} illustrates the superiority of our proposed AAD metric over the existing AD metric. In (b), our dewarped result is visibly superior to that of UVDoc~\citepalias{verhoeven2023uvdoc} due to better text-line alignment, although the two appear broadly similar. The AD metric, however, fails on two counts: its heatmap in (c) is uninterpretable, and its numerical score contradicts this visual reality by rating our result as worse.
In contrast, our AAD metric is founded on a clear geometric principle. This gives its heatmap in (d) a highly interpretable meaning, where brighter areas directly correspond to distorted feature lines. As a result, both its visual heatmap and numerical score correctly identify our result as superior, which is consistent with human judgment.
This ability to distinguish subtle differences is essential for guiding the next stage of research as performance gaps between state-of-the-art models narrow.

\begin{table*}[!htbp]
    \centering
    \footnotesize
    \caption{Quantitative evaluation and ablation study on the DocUNet and DIR300 benchmarks. We compare our method with previous methods and ablate our components: Axis-aligned geometric constraint Loss (AL) and Axis alignment Preprocessing (AP). “↑” indicates higher is better, “↓” means lower is better. \textbf{Bold} denotes the best result, and \underline{underline} indicates second-best. The “Improvements” row shows the relative gain of our full model over the previous state-of-the-art.}
    \label{tab:comparison_combined}
    \setlength{\tabcolsep}{4.5pt} 
    \begin{tabular}{l cccccc | cccccc}
        \toprule
        \multirow{2}{*}{\textbf{Method}} & \multicolumn{6}{c}{\textbf{DocUNet benchmark}} & \multicolumn{6}{c}{\textbf{DIR300 benchmark}} \\
        \cmidrule(lr){2-7} \cmidrule(lr){8-13}
        & \textbf{MS↑} & \textbf{LD↓} & \textbf{AD↓} & \textbf{AAD↓} & \textbf{ED↓} & \textbf{CER↓} & \textbf{MS↑} & \textbf{LD↓} & \textbf{AD↓} & \textbf{AAD↓} & \textbf{ED↓} & \textbf{CER↓} \\
        \midrule
        DewarpNet\cite{das2019dewarpnet}      & 0.474          & 8.362          & 0.398          & 0.164          & 824.5          & 0.225          & 0.492          & 13.944         & 0.332          & 0.147          & 1076.8         & 0.336 \\
        DDCP\cite{xie2021document}$^*$        & 0.472          & 8.982          & 0.428          & 0.159          & 1124.7        & 0.280          & 0.552          & 11.02          & 0.357          & 0.132          & 739.3          & 0.257 \\
        DocTr\cite{feng2021doctr}            & 0.509          & 7.773          & 0.369          & 0.151          & 708.6          & 0.185          & 0.616          & 7.189          & 0.255          & 0.107          & 698.4          & 0.211 \\
        PaperEdge\cite{ma2022learning}       & 0.473          & 7.967          & 0.368          & \underline{0.121} & 731.4          & 0.183          & 0.584          & 7.968          & 0.255          & 0.091          & \underline{447.4} & 0.177 \\
        DocGeoNet\cite{feng2022geometric}    & 0.504          & 7.717          & 0.381          & 0.158          & 695.0          & 0.187          & 0.638          & 6.403          & 0.241          & 0.106          & 672.5          & 0.206 \\
        FTDR\cite{li2023foreground}           & 0.497          & 8.416          & 0.377          & 0.151          & \underline{688.3} & \underline{0.176} & 0.607          & 7.680          & 0.243          & 0.108          & 639.4          & 0.198 \\
        LADoc\cite{li2023layout}$^\dagger$      & 0.525          & \underline{6.706} & \underline{0.300} & 0.121          & 689.8          & 0.180          & \underline{0.652} & \underline{5.702} & \underline{0.195} & \underline{0.087} & 495.4          & \underline{0.173} \\
        UVDoc\citepalias{verhoeven2023uvdoc} & \textbf{0.545} & 6.827          & 0.316          & 0.125          & 754.2          & 0.193          & 0.621          & 7.730          & 0.219          & 0.101          & 614.0          & 0.237 \\
        \midrule
        \quad Ours (Baseline, w/o AL, AP)     & 0.538          & 6.897          & 0.322          & 0.124          & 638.1          & 0.167          & 0.615          & 7.691          & 0.215          & 0.100          & 633.0          & 0.243 \\
        \quad Ours (+ AL only)                & 0.549          & 6.566          & 0.291          & 0.105          & 718.0          & 0.183          & 0.629          & 7.880          & 0.176          & 0.076          & 498.9          & 0.206 \\
        \quad Ours (+ AP only)                & 0.536          & 6.767          & 0.308          & 0.120          & 628.5          & 0.165          & 0.694          & 4.303          & 0.152          & 0.069          & 541.6          & 0.153 \\
        \textbf{Ours (Full, AL+AP)}           & \underline{0.543} & \textbf{6.249} & \textbf{0.278} & \textbf{0.099} & \textbf{603.1} & \textbf{0.150} & \textbf{0.702} & \textbf{4.261} & \textbf{0.131} & \textbf{0.057} & \textbf{405.8} & \textbf{0.132} \\
        \midrule
        \textbf{Improvements}                 & -0.4\%         & 6.8\%          & 7.3\%          & 18.2\%         & 12.4\%         & 14.8\%         & 7.7\%          & 25.3\%         & 32.8\%         & 34.5\%         & 9.3\%          & 23.7\% \\
        \bottomrule
    \end{tabular}
    
    \begin{flushleft}
    \footnotesize 
    \textit{Note:} For a fair OCR metrics comparison, all evaluations are performed at the original input resolution. The dewarped images from DDCP $^*$ are upscaled and those from LADoc $^\dagger$ are downscaled to match this resolution before metric calculation.
    \end{flushleft}
\end{table*}

\subsection{Implementation Details}
Our model is implemented using the PyTorch framework~\cite{paszke2017automatic}. The dewarping model is trained on two datasets: Doc3D~\cite{das2019dewarpnet} (88k images) and UVDoc~\citepalias{verhoeven2023uvdoc} (20k images). The input images are resized to 712$\times$488 pixels, while the output 2D grid is set to 45$\times$31. We optimize the model using the AdamW optimizer~\cite{loshchilov2017decoupled} with a batch size of 72, with an initial learning rate of $1\times10^{-5}$ and a linear decay schedule. The network is trained for 30 epochs on two NVIDIA 4090 GPUs, which takes approximately 15 hours.
During inference (see~\Cref{sec:rotation_alignment}), we apply preprocessing once for the DocUNet benchmark and twice for DIR300 to better handle its smaller target objects.

\subsection{Experimental Results}
We quantitatively and qualitatively evaluate our method against others on the DocUNet and DIR300 benchmarks.

\Cref{tab:comparison_combined} summarizes the quantitative evaluation. On the DocUNet benchmark, our approach demonstrates significant improvements across most metrics, achieving performance comparable to UVDoc~\citepalias{verhoeven2023uvdoc} on the MS metric and delivering a notable 18.2\% improvement on the proposed AAD metric. The improvements are even more pronounced on the larger DIR300 benchmark, where our method achieves gains across all metrics, including a 34.5\% improvement in AAD and a significant boost to OCR performance, with the CER reaching 23.7\%.

\begin{figure*}[!ht]
    \centering
    \includegraphics[width=0.995\linewidth]{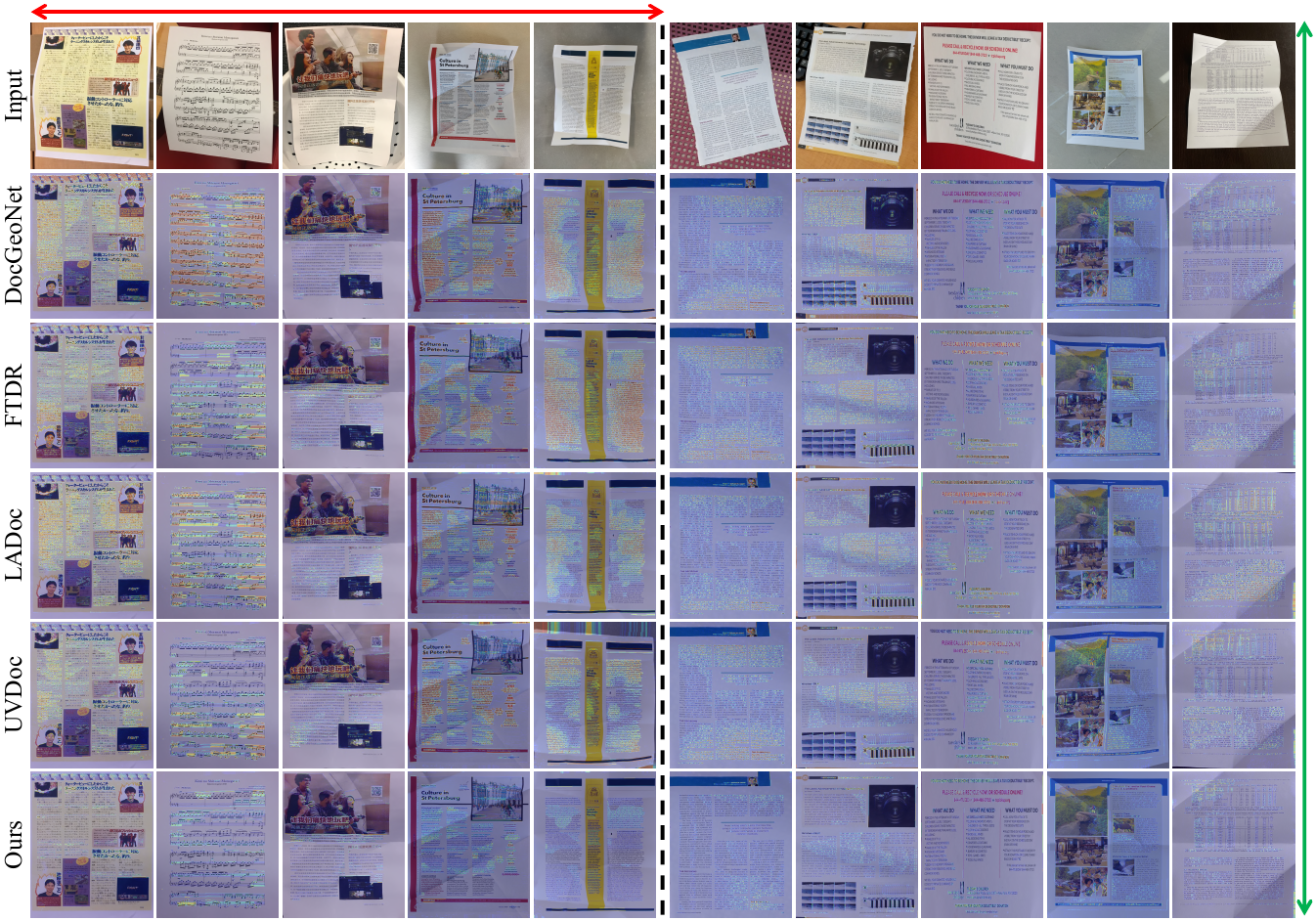}
    \caption{Qualitative evaluation. The figure shows the dewarped results along with the corresponding heatmap overlays of the AAD component metric. The horizontal ($\leftrightarrow$) and vertical ($\updownarrow$) arrows indicate the directions for visualizing the axis-aligned error of the dewarped feature lines.}
    \label{fig:compare_img_docunet_dir300}
\end{figure*}

Qualitative results for both benchmarks are shown in \Cref{fig:compare_img_docunet_dir300}. To highlight distortion areas more clearly, we overlay the dewarped results with heatmaps generated from the horizontal and vertical components of our AAD metric (an example is provided in \Cref{fig:AAD}). For our method, the resulting heatmaps appear consistently fainter on documents from both datasets, indicating substantially reduced axis-alignment error. As expected from the dewarping results in \Cref{fig:intro_img}(a), this yields more axis-aligned feature lines in the final dewarped images.

\begin{figure}[!htbp]
    \centering
    \includegraphics[width=1.0\linewidth]{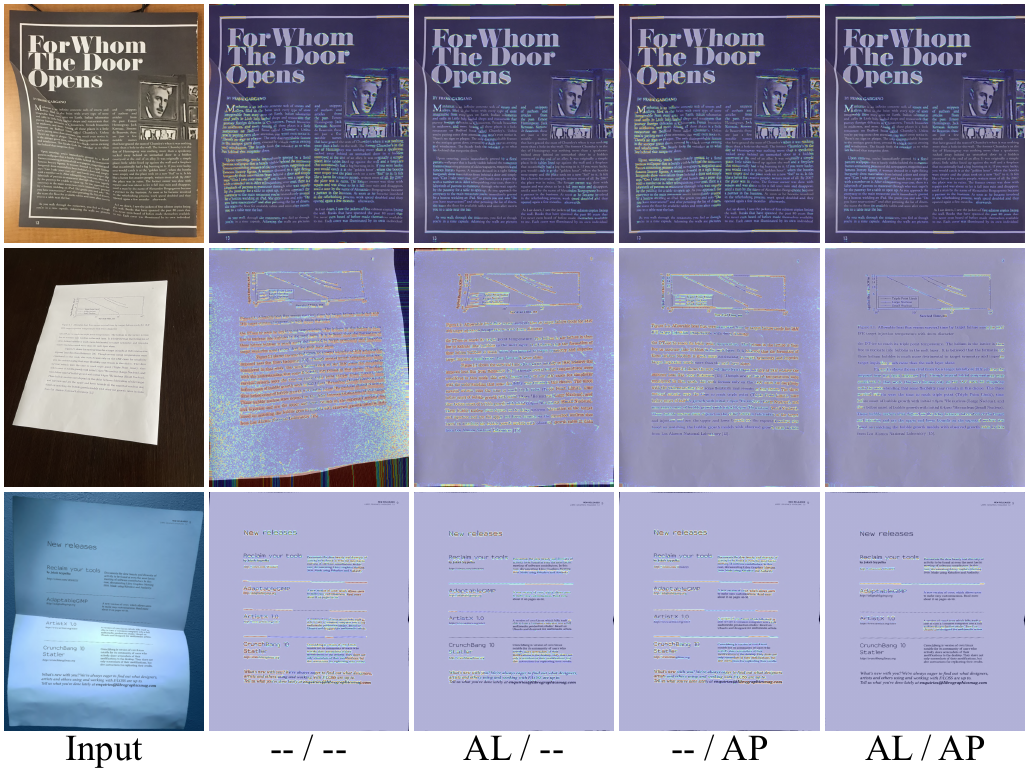}
    \caption{
    Qualitative results of our dewarping ablation study. From left to right: the distorted input and dewarped outputs from different settings. The overlaid heatmaps visualize the horizontal component of the AAD metric.}
    \label{fig:ablation}
\end{figure}

\subsection{Ablation Study}\label{sec:ablation study}

We conducted ablation experiments to evaluate the impact of two key components: the axis-aligned geometric constraint used during training (\Cref{sec:slgc}) and the axis alignment preprocessing applied during inference(\Cref{sec:rotation_alignment}). 

As summarized in~\Cref{tab:comparison_combined}, both components independently enhance performance across two benchmark datasets. Specifically, for the DocUNet benchmark—where a large proportion of targets are present—the axis-aligned geometric constraint yields greater performance improvements than the axis alignment preprocessing. Conversely, in the DIR300 benchmark, which contains a smaller proportion of targets, the benefits of the preprocessing strategy are more pronounced. Importantly, the combination of both modules produces the best overall performance, demonstrating their complementary advantages in different scenarios.

\Cref{fig:ablation} provides several examples from our ablation experiments. For targets with a smaller proportion (middle row), the preprocessing strategy improves target integrity, while the addition of the axis-aligned geometric constraint further enhances the dewarping performance. For targets with a larger proportion (first and last rows), the performance gains from the axis-aligned geometric constraint are more prominent. Overall, the optimal dewarped results are achieved by applying both components simultaneously, which aligns with the improvements observed in the qualitative metrics in~\Cref{tab:comparison_combined}.

\section{Conclusion} \label{sec:Conclusion}
This paper proposes a document dewarping method that exploits the inherent axis-aligned property of planar documents' underlying grid geometry. To this end, we introduce two key components: an axis-aligned geometric constraint for training and an alignment preprocessing strategy for inference. Experimental results confirm that these components are both independently effective and complementary. As a result, our full model outperforms state-of-the-art methods, demonstrating both significant quantitative improvements and clear qualitative advantages. To apply our geometric insight to the evaluation process itself, we introduce a novel evaluation metric AAD, which provides a clear geometric interpretation, aligns well with human visual perception, and exhibits enhanced robustness.
Unlike previous work focused on network architectures or datasets, our approach leverages the intrinsic geometry of documents to improve dewarping performance. This principle-driven strategy demonstrates the significant value of exploiting such properties. Future work will involve extending this principle to other image rectification tasks.

\section{Acknowledgments}
This work was supported by the National Natural Science Foundation of China (Grant No. 62495092), the Natural Science Basic Research Plan of Shaanxi Province, China (No. 2025SYS-SYSZD-023), and the China Scholarship Council (No. 202406280284).

\bibliography{aaai2026}

\clearpage
\twocolumn[\begin{@twocolumnfalse}
  \begin{center}
    {\huge \textbf{Supplementary Material}}
  \end{center}
  \vspace{1.5em}
\end{@twocolumnfalse}]
\appendix
\setcounter{page}{1}
\setcounter{table}{0}  
\setcounter{figure}{0} 
\renewcommand{\thetable}{S\arabic{table}}  
\renewcommand{\thefigure}{S\arabic{figure}} 

\section*{OVERVIEW}
In this supplementary material, we provide the following additional information: \begin{enumerate}[label=(\arabic*)] 
\setlength{\itemindent}{5mm} 
\item Details of the Inference Process.
\item Details of the evaluation metrics. 
\item Robustness analysis of the AAD Metric.
\item Comparison on UVDoc Benchmark.
\item Additional visual comparison of the AD and AAD metrics.
\item Visual comparison of OCR results on dewarped document images. 
\item Additional experimental results. 
\end{enumerate}

\section{Details of the Inference Process}
During inference, we first rotate and crop the document region using the axis alignment preprocessing method described in the main text. To ensure that the entire document area is preserved, we employ a region expansion strategy instead of merely clipping the edges based on the minimum rotation matrix boundary. 

The expansion ratio is adjusted for each benchmark: for the UVDoc benchmark~\cite{verhoeven2023uvdoc}, it is set to 0.03; for the DIR300 benchmark~\cite{feng2022geometric}, an initial expansion of 0.01 is followed by a secondary expansion of 0.03; and for the DocUNet benchmark~\cite{ma2018docunet}, the ratio is set to 0.06.

\section{Details of the Evaluation Metrics}
The multi-scale structural similarity (MS-SSIM) metric extends SSIM by evaluating image similarity at multiple scales and combining these measurements through a weighted average. We adopt the original weights from~\cite{wang2003multiscale}: [0.0448, 0.2856, 0.3001, 0.2363, 0.1333].

Local Distortion (LD) is computed using dense SIFT flow mapping~\cite{liu2010sift} from the ground truth image to the dewarped image, and is defined as the average L2 distance between corresponding pixels~\cite{you2017multiview}. This metric effectively measures the average local distortion of the dewarped image relative to the ground truth.
Aligned Distortion (AD) is a more robust variant introduced in~\cite{ma2022learning}. Unlike LD, AD mitigates errors caused by global image translation and scaling by extracting the optimal affine transformation from the SIFT flow. It further weights the error according to the image gradient magnitude, emphasizing regions with significant features such as text or edges.
In addition, the AAD metric proposed in the main text offers enhanced geometric interpretability and robustness compared to AD. We compute SIFT flow using the official implementation in MATLAB 2024b. Prior to computing these metrics, all dewarped and reference images are resized to a total area of 598,400 pixels, as recommended in~\cite{ma2018docunet}.

Beyond image similarity metrics, we also evaluate OCR performance using character error rate (CER) and edit distance (ED)~\cite{navarro2001guided}. Edit distance quantifies the difference between two character sequences, while CER is defined as the ratio of the edit distance to the total number of characters in the reference text, i.e., (s + i + d)/N, where s, i, and d denote the number of substitutions, insertions, and deletions, respectively, and N is the total number of characters in the reference text.

\section{Robustness Analysis of the AAD Metric}
\begin{figure}[!htbp]
    \centering
    \includegraphics[width=1.0\linewidth]{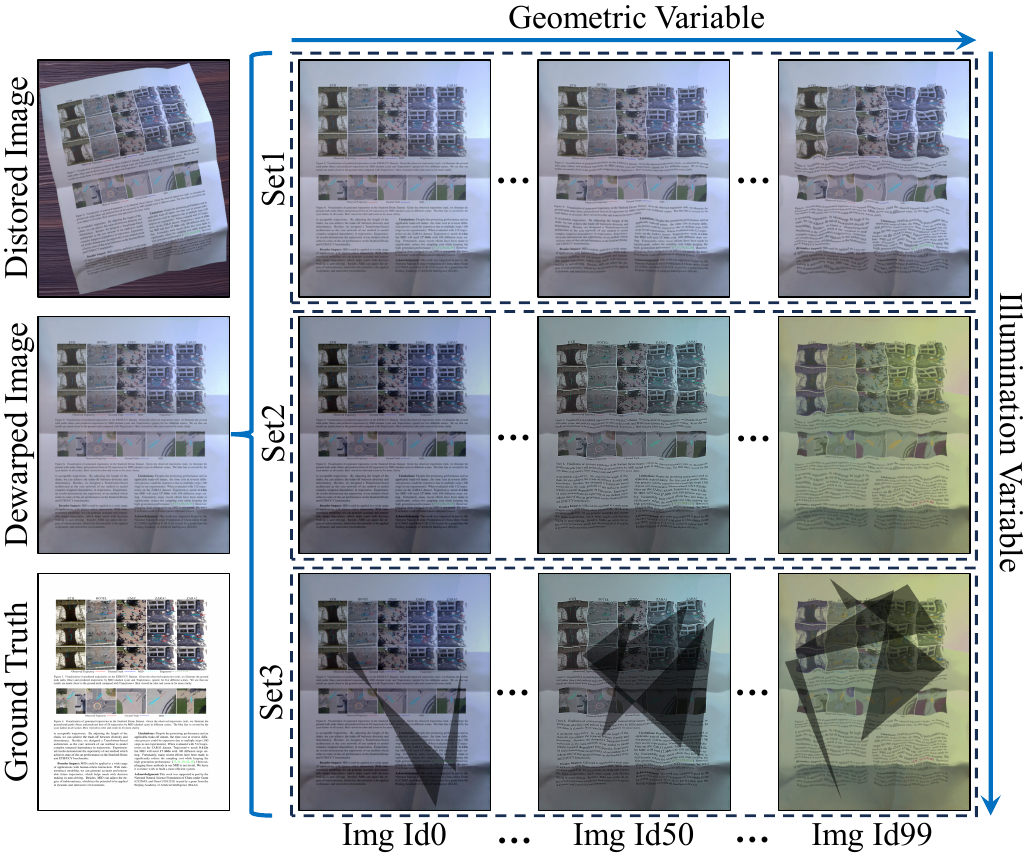}
  \caption{Geometric and illumination disturbances applied to the dewarped image. The leftmost column shows the input distorted image, alongside its corresponding dewarped and ground truth versions. On the right, geometric and illumination disturbances are added to the dewarped image, and the intensity increases along with the corresponding arrow direction. Set1 includes only geometric disturbances, Set2 adds color interference to Set1, and Set3 adds a random shadow to Set2.}
    \label{fig:AAD_generateimg}
\end{figure}

To validate the robustness of the proposed AAD metric, we applied geometric and illumination disturbances to the dewarped images shown in~\Cref{fig:AAD_generateimg}, generating 100 warped images for each setting. 
To create a ground truth for evaluation, we calculate the 'true' metric values using the known geometric disturbances as the ground truth flow. We then compute the 'reference' metric values using the flow field estimated by the SIFT-flow~\cite{liu2010sift} method. The correlation between these true and reference values indicates the metric's robustness.

\begin{figure}[!htbp]
    \centering
    \includegraphics[width=1.0\linewidth]{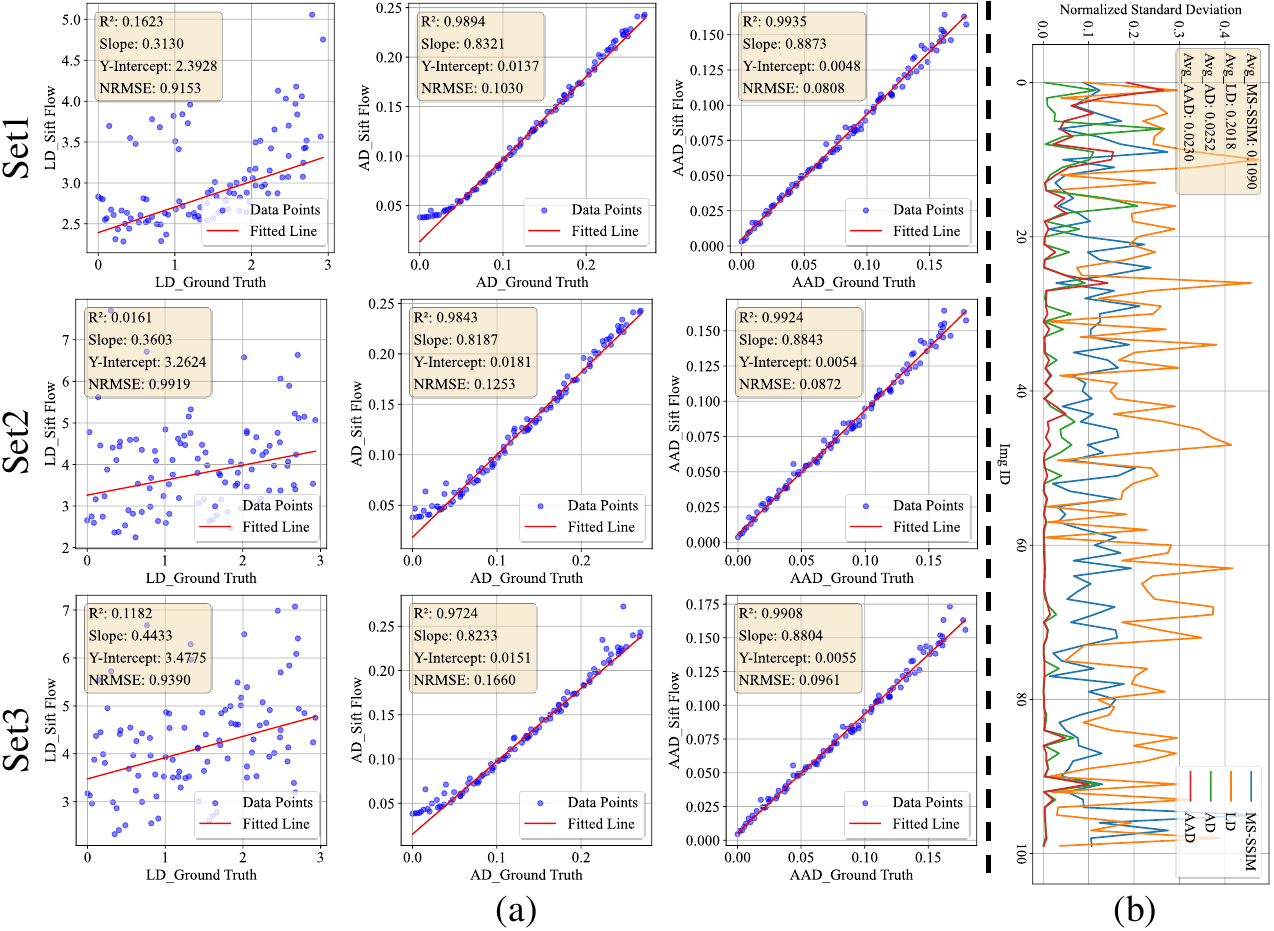}
     \caption{Robustness analysis of each metric. (a) The x-axis represents the ground truth metric computed from the ground truth optical flow, while the y-axis represents the reference metric derived from the SIFT-flow method; ideally, they should exhibit a high linear correlation. (b) The normalized standard deviation of the metrics under three different environmental conditions, lower values indicate greater robustness against environmental disturbances.}
    \label{fig:AAD_static}
\end{figure}

As shown in~\Cref{fig:AAD_static}(a), the AAD metric demonstrated the highest linear correlation (\(R^{\text{2}}\)) across all sets, indicating a strong linear relationship with the ground truth even under minimal geometric disturbances(i.e., in regions with lower metric values, which are critical for comparing dewarped results that closely approximate the true values). In contrast, the AD metric exhibited a nonlinear response in these regions.

In~\Cref{fig:AAD_static}(b), we evaluated the normalized standard deviation under three different illumination conditions for each image in~\Cref{fig:AAD_generateimg}. The results revealed that the standard deviation for AAD (0.0230) was smaller than that for AD (0.0252), further confirming its enhanced robustness.

\section{Comparison on UVDoc benchmark}
This section provides a detailed analysis of our experimental results on the UVDoc benchmark~\citepalias{verhoeven2023uvdoc}. UVDoc is a synthetic dataset generated via a pseudo-photorealistic rendering pipeline. This benchmark consists of 50 images, including matching pairs with and without illumination. Following the UVDoc protocol, 20 rich-text images were selected for OCR evaluation.

\begin{figure*}[!htbp]
    \centering
    \includegraphics[width=1.0\linewidth]{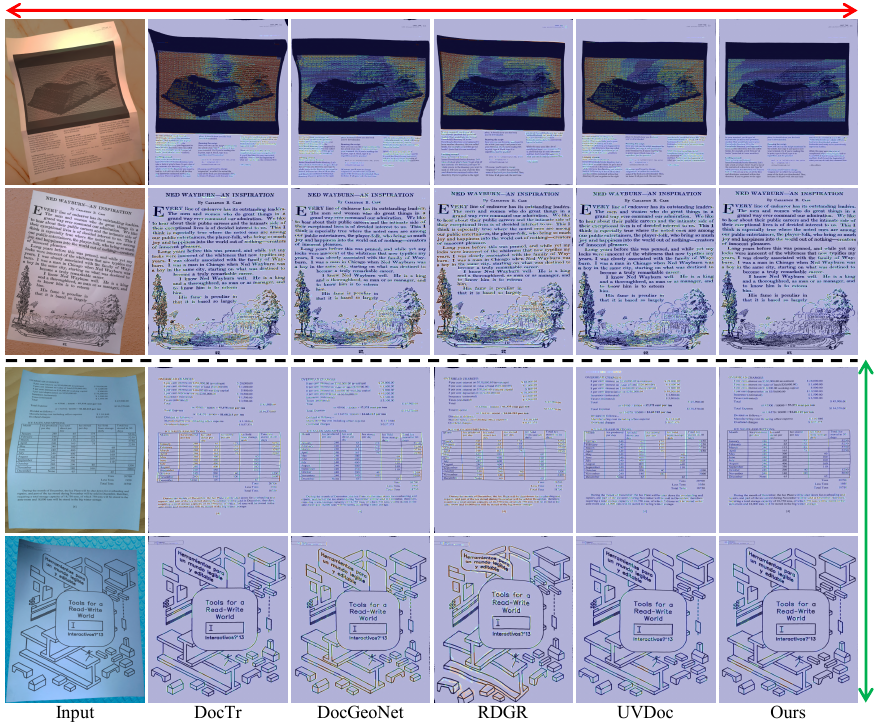}
    \caption{Qualitative comparisons on the UVDoc benchmark. The figure shows the dewarped results along with the corresponding heatmap overlays of the AAD component metric. The horizontal ($\leftrightarrow$) and vertical ($\updownarrow$) arrows indicate the directions for visualizing the axis-aligned error of the dewarped feature lines.}
    \label{fig:compare_img_uvdoc}
\end{figure*}

In~\Cref{tab:comparison_uvdoc}, we present the quantitative results on the UVDoc benchmark. Our method outperforms competing approaches across all metrics except for LD and CER, where our results are very close to the best values. Notably, the AAD metric improves by 23.1\%.
The qualitative comparison in \Cref{fig:compare_img_uvdoc} displays dewarped results for both horizontal and vertical directions.
These comparisons reveal that our method achieves excellent axis-aligned dewarping performance even in regions without textual features, highlighting the superiority of our approach.

The corresponding ablation results in~\Cref{tab:comparison_uvdoc} are consistent with the other two benchmarks in the main text. The two proposed modules are individually effective, and they can be combined to effectively enhance performance.

\section{Additional Visual Comparison of the AD and AAD metrics}
\begin{figure*}[!htbp]
    \centering
    \includegraphics[width=1.0\linewidth]{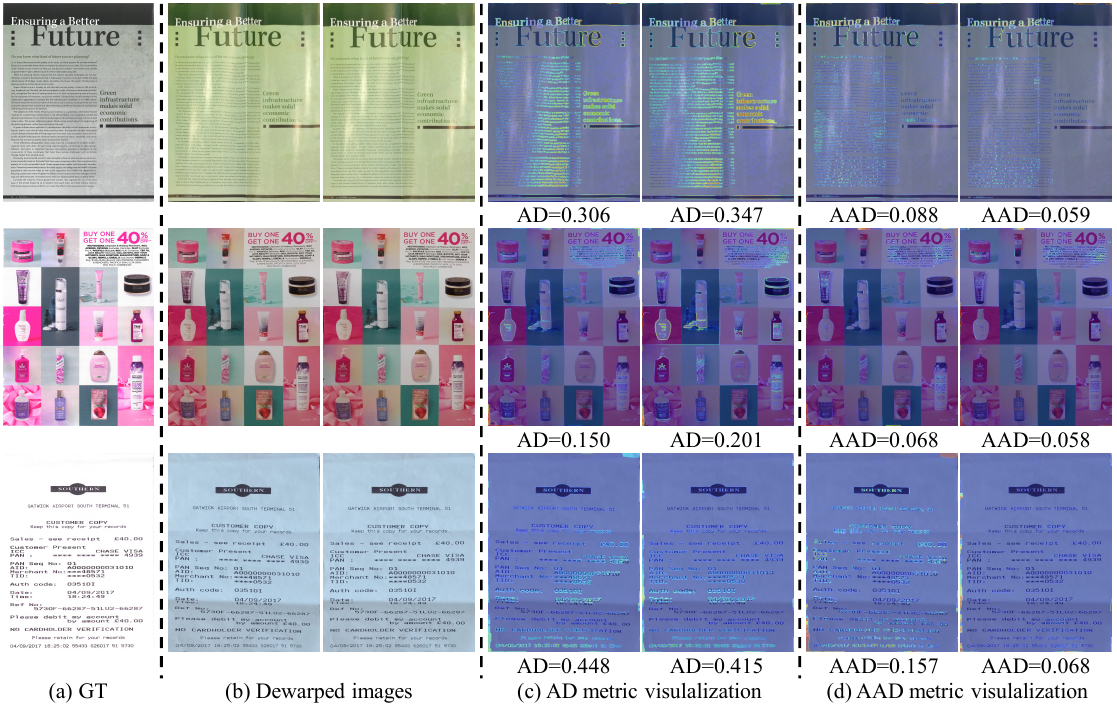}
    \caption{Visualization of the AD and AAD metrics. (a) Ground truth image. (b) Comparison of dewarping results: the left image is produced by UVDoc~\citepalias{verhoeven2023uvdoc} and the right image is generated by our method. (c) Heatmap of the AD metric computed for the dewarped images relative to the ground truth. (d) Heatmap corresponding to the AAD metric proposed in the main text.}
    \label{fig:supply_compare_ad_aad}
\end{figure*}

In~\Cref{fig:supply_compare_ad_aad}, we have provided additional case comparisons between the AAD and the AD metric. Each row contrasts the dewarped results obtained using UVDoc~\citepalias{verhoeven2023uvdoc} with those produced by our method, accompanied by corresponding heatmap visualizations for each metric. Notably, the differences observed in~\Cref{fig:supply_compare_ad_aad}(b) are primarily due to the axis alignment of text lines. These differences are distinctly captured by the AAD heatmap in~\Cref{fig:supply_compare_ad_aad}(d), reinforcing that the AAD metric more closely aligns with human perception of dewarping quality.

In contrast, the AD heatmap in~\Cref{fig:supply_compare_ad_aad}(c) lacks such discriminative features. Its color variations do not convey clear, meaningful distinctions, and the numerical values indicate that AD fails to effectively differentiate subtle differences. For example, the AD metric assigns a lower (better) score to the result on the left than the one on the right, a conclusion that contradicts visual assessment.

Our proposed AAD metric, however, successfully captures these subtle differences. When dewarping results deviate only slightly from the ground truth, the AD metric struggles to capture these nuances and does not maintain a consistent linear relationship with the true values—a limitation that could hinder future method improvements. In contrast, the AAD metric is more robust, interpretable, and closely aligned with human perception, making it an excellent evaluation metric for dewarping tasks.

\section{Visual Comparison of OCR Results on Dewarped Document Images}
\begin{figure*}[!htbp]
    \centering
    \includegraphics[width=1.0\linewidth]{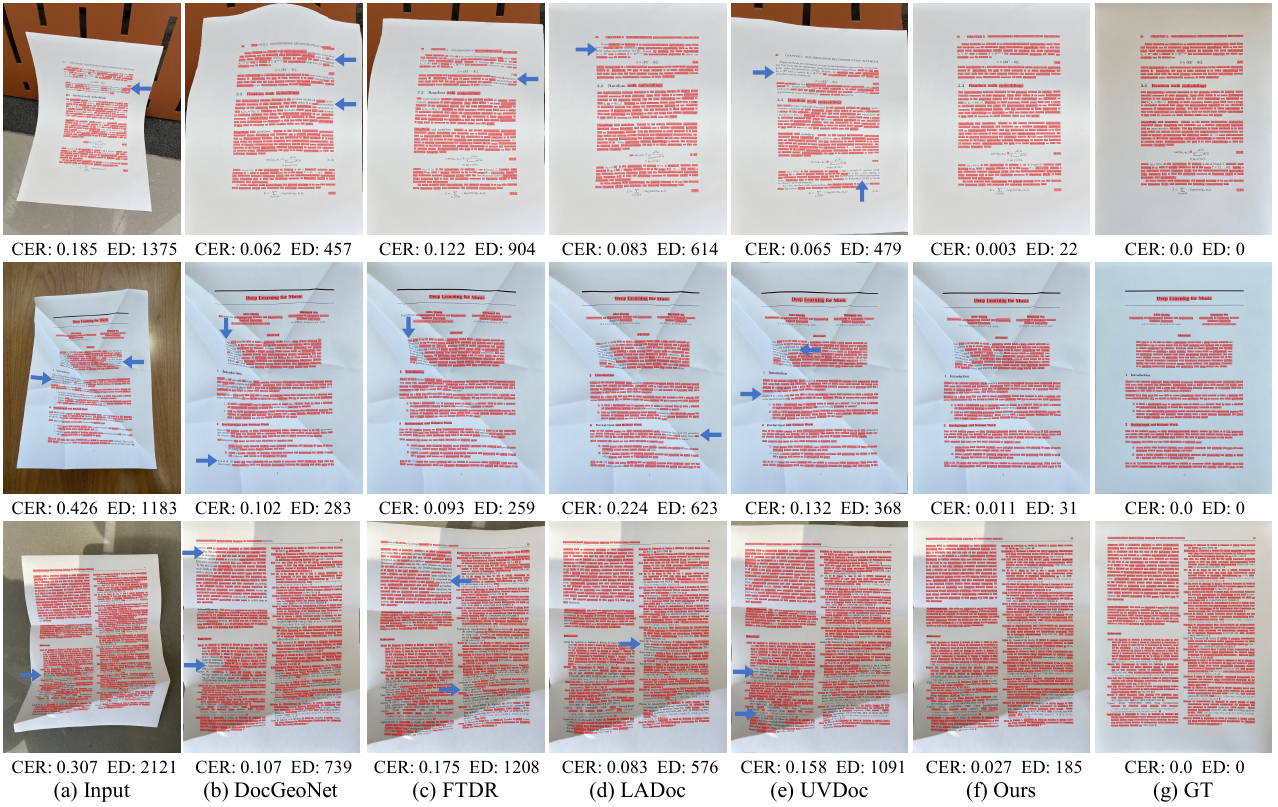}
    \caption{OCR results for dewarped documents produced by various methods. Each row overlays the detected text regions on the dewarped image derived from the distorted input (first column). Red boxes indicate recognized text areas, while blue arrows mark regions where text was missed. The number below each image represents the OCR metrics relative to the ground truth (GT) recognition result shown in the last column.}
    \label{fig:supply_compare_ocr}
\end{figure*}

\Cref{fig:supply_compare_ocr} illustrates several cases demonstrating how dewarping results from different methods affect OCR performance. In each instance, we overlay the detected character bounding boxes on the dewarped images. Our method reliably recognizes text in nearly all regions, outperforming alternative approaches. Moreover, the minimal overlap between adjacent text boxes underscores the superior correction quality, yielding outputs that closely resemble the ground truth. Below each image is the computed OCR metric value, which shows a significant reduction in errors compared to the distorted input. These results confirm that improved dewarping directly contributes to lower OCR error rates, bringing the output closer to the ideal ground truth.

\section{Additional Experimental Results}
\begin{figure*}[!htbp]
    \centering
    \includegraphics[width=1.0\linewidth]{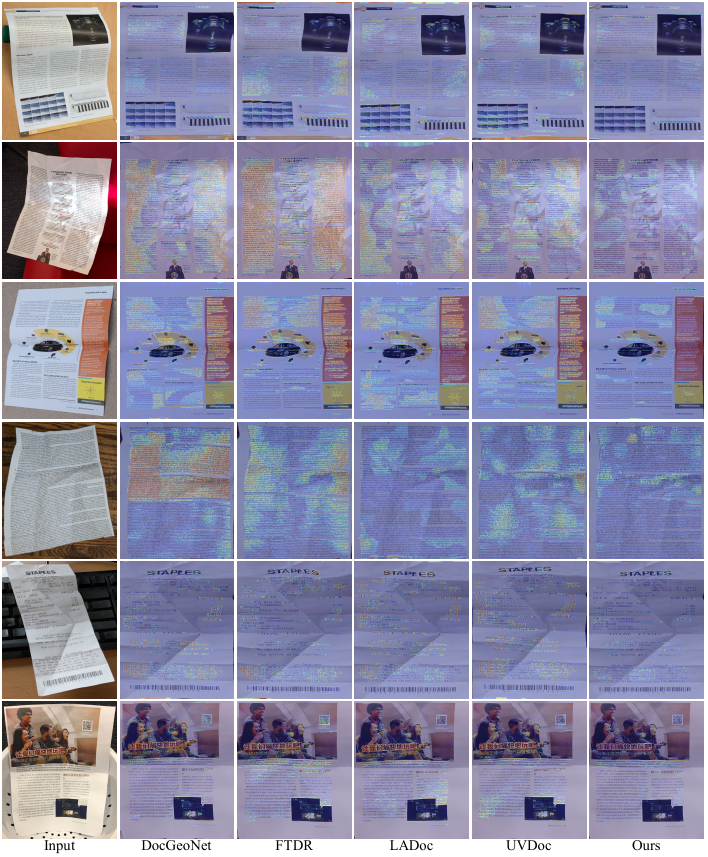}
    \caption{Qualitative evaluation on the DocUNet benchmark. Each panel shows the dewarped result with the corresponding horizontal AAD metric heatmap overlaid, allowing for a detailed comparison of correction errors among different methods.}
    \label{fig:supply_compare_img_docunet}
\end{figure*}

\begin{figure*}[!htbp]
    \centering
    \includegraphics[width=1.0\linewidth]{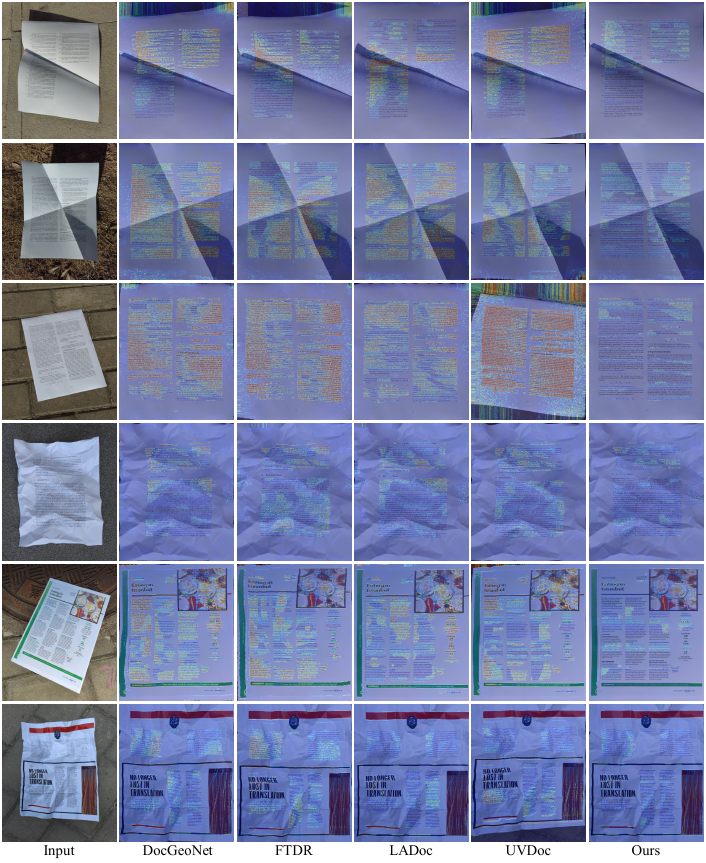}
    \caption{Qualitative evaluation on the DIR300 benchmark. Each panel shows the dewarped result with the corresponding horizontal AAD metric heatmap overlaid, allowing for a detailed comparison of correction errors among different methods.}
    \label{fig:supply_compare_img_DIR300}
\end{figure*}

\begin{figure*}[!htbp]
    \centering
    \includegraphics[width=1.0\linewidth]{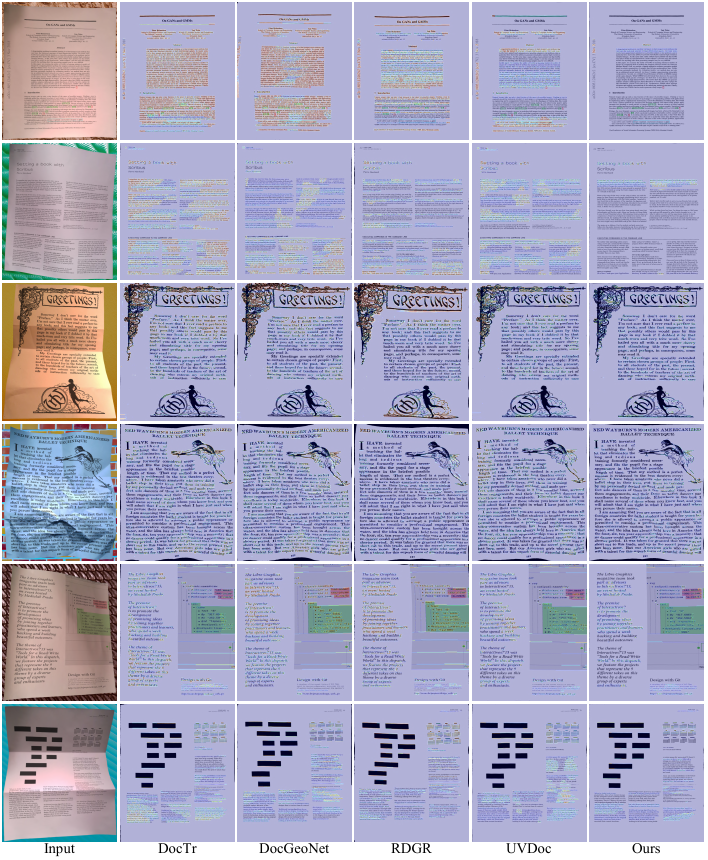}
    \caption{Qualitative evaluation on the UVDoc benchmark. Each panel shows the dewarped result with the corresponding horizontal AAD metric heatmap overlaid, allowing for a detailed comparison of correction errors among different methods.}
    \label{fig:supply_compare_img_UVDOC}
\end{figure*}

Owing to space limitations in the main text, we include additional dewarping results in this supplementary material.~\Cref{fig:supply_compare_img_docunet},~\Cref{fig:supply_compare_img_DIR300}, and \Cref{fig:supply_compare_img_UVDOC} present sample outputs on the DocUNet~\cite{ma2018docunet}, DIR300~\cite{feng2022geometric}, and UVDoc~\citepalias{verhoeven2023uvdoc} benchmarks, respectively. These examples span a diverse range of document layouts. To facilitate a detailed assessment of the correction accuracy, we overlay the horizontal component heatmap of our AAD metric onto the dewarped images. As shown in the figures, our method consistently yields less distorted results with fainter heatmap intensities—indicating lower correction errors—compared to other approaches. These additional results further validate the robustness and superiority of our proposed method.

\begin{table*}[!htbp]
    \centering
    \footnotesize
    \caption{Quantitative evaluation and ablation study on the UVDoc benchmark. We compare our method with previous methods and ablate our components: Axis-aligned geometric constraint Loss (AL) and Axis alignment Preprocessing (AP). “↑” indicates higher is better, “↓” means lower is better. \textbf{Bold} denotes the best result, and \underline{underline} indicates second-best. The “Improvements” row shows the relative gain of our full model over the previous state-of-the-art.}
    \label{tab:comparison_uvdoc}
    \setlength{\tabcolsep}{5.5pt}
    \begin{tabular}{lcccccc}
        \toprule
        \textbf{Method} & \textbf{MS↑} & \textbf{LD↓} & \textbf{AD↓} & \textbf{AAD↓} & \textbf{ED↓} & \textbf{CER↓} \\
        \midrule
        DewarpNet\cite{das2019dewarpnet} & 0.589 & \textbf{5.666} & 0.193 & 0.085 & 297.8 & 0.117 \\
        DDCP\cite{xie2021document} & 0.585 & 11.582 & 0.290 & 0.109 & 251.8 & 0.094 \\
        DocTr\cite{feng2021doctr} & 0.697 & 8.146 & 0.161 & 0.075 & \underline{110.8} & \textbf{0.034} \\
        DocGeoNet\cite{feng2022geometric} & 0.706 & 8.350 & 0.168 & 0.072 & 123.1 & 0.047 \\
        RDGR\cite{jiang2022revisiting} & 0.610 & 6.423 & 0.281 & 0.096 & 122.45 & 0.051 \\
        UVDoc\citepalias{verhoeven2023uvdoc} & \underline{0.785} & 6.810 & \underline{0.119} & \underline{0.052} & 122.7 & 0.046 \\
        \midrule
        \quad Ours (Baseline, w/o AL, AP) & 0.794 & 6.356 & 0.118 & 0.052 & 120.4 & 0.041 \\
        \quad Ours (+ AL only) & 0.785 & 5.903 & 0.116 & 0.044 & 105.2 & 0.035 \\
        \quad Ours (+ AP only) & 0.790 & 6.024 & 0.111 & 0.049 & 118.2 & 0.043 \\
        \textbf{Ours (Full, AL+AP)} & \textbf{0.800} & \underline{5.834} & \textbf{0.098} & \textbf{0.040} & \textbf{104.3} & \underline{0.036} \\
        \midrule
        \textbf{Improvements} & +1.9\% & -3.0\% & +17.6\% & +23.1\% & +5.9\% & -5.9\% \\
        \bottomrule
    \end{tabular}
\end{table*}

\end{document}